\documentclass{article}
\usepackage[preprint]{spconf,amsmath,graphicx}
\usepackage{hyperref}
\usepackage{multirow}
\usepackage{rotating}

\newcommand\blfootnote[1]{%
  \begingroup
  \renewcommand\thefootnote{}\footnote{#1}%
  \addtocounter{footnote}{-1}%
  \endgroup
}

\title{A STUDY ON CROSS-CORPUS SPEECH EMOTION RECOGNITION AND DATA AUGMENTATION}
\name{Norbert Braunschweiler, Rama Doddipatla, Simon Keizer, Svetlana Stoyanchev}
\address{Toshiba Europe Limited, Cambridge Research Laboratory, Cambridge, CB4 0GZ, UK}

\begin{document}
\ninept
\maketitle
\begin{abstract}
Models that can handle a wide range of speakers and acoustic conditions
are essential in speech emotion recognition (SER). Often, these models tend
to show mixed results when presented with speakers or acoustic conditions
that were not visible during training. This paper investigates the impact
of cross-corpus data complementation and data augmentation on the
performance of SER models in matched (test-set from same corpus) and 
mismatched (test-set from different corpus) conditions.
Investigations using six emotional speech corpora that include
single and multiple speakers as well as variations in emotion
style (acted, elicited, natural) and recording conditions are presented.
Observations show that, as expected, models trained on single corpora perform best
in matched conditions while performance decreases between 10-40\%
in mismatched conditions, depending on corpus specific features.
Models trained on mixed corpora can be more stable in mismatched contexts, 
and the performance reductions range from 1 to 8\% when compared with single corpus
models in matched conditions. Data augmentation yields additional
gains up to 4\% and seem to benefit mismatched conditions more than matched ones.
\end{abstract}

 \vspace{-0.2cm}
\section{INTRODUCTION}
\label{sec:intro}

Emotion recognition is attracting wide research interest in a variety of areas
of which the most dominant one aims to improve human-machine interaction, for example,
to generate more adequate responses in content and style for dialogue systems.
Moreover, it is used in other domains such as supporting doctors
to tailor therapeutic techniques for patients 
or in the recording and analysis of emotional states
during meetings and interviews. 
While there has been a lot of progress in emotion recognition, especially in
speech emotion recognition (SER), there are still unresolved challenges to improve
the performance as well as the robustness of models towards speakers and conditions not
seen in training. Deep learning models may perform reasonably well on a single corpus,
but often exhibit significantly lower performance when
faced with new speakers in different acoustic environments. 
\blfootnote{Copyright 2021 IEEE. Published in the 2021 IEEE Automatic Speech Recognition and Understanding Workshop (ASRU) (ASRU 2021), scheduled for 14-18 December 2021 in Cartagena, Colombia. Personal use of this material is permitted. However, permission to reprint/republish this material for advertising or promotional purposes or for creating new collective works for resale or redistribution to servers or lists, or to reuse any copyrighted component of this work in other works, must be obtained from the IEEE. Contact: Manager, Copyrights and Permissions / IEEE Service Center / 445 Hoes Lane / P.O. Box 1331 / Piscataway, NJ 08855-1331, USA. Telephone: + Intl. 908-562-3966.}

Approaches to increase the performance of SER models on individual corpora
are plentiful in the literature, 
but the situation of handling a broad range of speaking styles and acoustic scenes
is still challenging. Also, when focusing on single corpora there is no need to
consider the details of different annotation schemes as mentioned
in \cite{Atcheson2018_AnnotatorDeisagreement}.

In contrast, the area of cross-corpus speech emotion recognition, in which
data from multiple corpora is used to train models, is looking into the aspects
of how multiple and often diverse corpora can help to improve SER performance
in general and robustness against unseen data in particular. The diversity of
corpora ranges from features such as language (cross-lingual),
emotion style (acted, elicited, natural) to acoustic scene (studio recording, regular room
recording, recordings with variable environments and background noises).

One of the questions in cross-corpus SER is whether the combination of corpora can improve
or degrade the performance.
The paper addresses this question on a selection of six emotional
speech corpora with a broad range of attributes including multi-speaker and single speaker corpora;
acted, elicited and what is deemed to be natural emotions; acoustic scenes ranging
from high quality studio recordings up to mixed quality recordings with variable background
noises. 

Using these diverse corpora a state-of-the-art CNN-RNN emotion
recognition model is trained and then tested in matched and mismatched conditions.
Furthermore, corpora are combined to measure the impact of cross-corpus complementation
on emotion classification performance in matched and mismatched conditions. 
Results are analyzed considering corpus specific characteristics
and more detailed comparisons are conducted for features such as scripted 
vs.\ improvised emotions and studio recordings vs.\ regular office recordings.

The paper presents related work next, followed by the description of the deep 
learning model and the augmentation methods. Then,
the individual corpora and their features are presented; the experimental set-up is laid out
and finally results are presented and discussed.

 \vspace{-0.2cm}
 \subsection{Related work}
\label{sec:RelatedWork} 
There are previous studies related to cross-corpus emotion recognition from
\cite{Schuller2010_CrossCrop} who use six corpora and different
types of normalization to handle variations observed across these corpora.
\cite{Liu_2018_DAlearing} introduce a domain-adaptive subspace learning method
to reduce the feature space differences between source and target speech.
In \cite{Milner2019_CROSSCORP} a bi-directional LSTM with attention mechanism
is used for classifying emotions across various corpora. To generalize
to emotions across corpora authors suggest to use models trained on
out-of-domain data and conduct adaptation to the missing corpus or
use domain adversarial training (DAT) \cite{Abdelwahab2018_DAT}.
\cite{Su_2021_GAN} use a conditional cycle emotion generative adversarial network
to generate synthetic data from the unlabeled target corpus to increase the
variability in the source corpus and subsequently improve performance in
cross-corpus  emotion recognition.
\cite{Lee_2021_Triplet} present an approach aimed at learning more
generalized features of emotional speech which uses a triplet network. 
Most recently \cite{ Zhang_2021_JDAR} proposed a subspace learning method called
joint distribution adaptive regression (JDAR) to reduce the feature distribution differences
between samples from the training and test sets.

Many of these works are based on relatively small emotional speech corpora
or mix different languages. While the introduction of larger corpora such as
CMU-MOSEI \cite{zadeh2018_MOSEI} which are also aimed at 
including more natural emotions has increased the amount of training data,
it is still interesting to investigate whether combining different corpora
can improve recognition results on other corpora as mentioned by
\cite{Milner2019_CROSSCORP}. 
The current study adds 3 large single speaker corpora into the pool
of available corpora for cross-corpus experiments thus enabling the study
of speaker individual influences in combination with multi-speaker
corpora of variable sizes and variable emotion types. This, in combination with
a new state-of-the-art deep learning model architecture provides
a fertile ground for interesting experiments. 

\vspace{-0.3cm}
\section{METHOD}
\label{sec:Method}
To investigate the impact of cross-corpus complementation and data augmentation,
first, a state-of-the-art deep learning model was established.
Then, focusing on the 4 emotion classes {\it angry}, {\it happy}, {\it sad}, {\it neutral},
which are the ones most widely found in emotional speech corpora,
corpus-specific models were trained and evaluated in matched and mismatched
conditions, followed by the creation of cross-corpus models and
a combination with data augmentation.
\vspace{-0.3cm}
\subsection{Model architecture}
\label{ssec:ModelArchitecture}
\begin{figure}[htb]
\begin{minipage}[b]{1.0\linewidth}
\centering
 \centerline{\includegraphics[width=8.3cm]{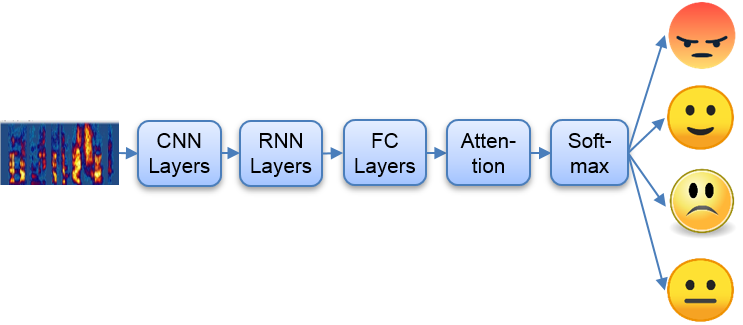}}
  \vspace{-0.4cm}
\end{minipage}
\caption{Illustration of the CNNRNNATT model architecture}
\label{fig:CNNRNNmodel}
\end{figure}

Inspired by the emotion recognition model presented in \cite{Jalal2020_BLSTMATT}
who derived their model from the triple attention network described in \cite{Beard_2018_TripleAtt},
who in turn based their model on work by \cite{Nam_2017}, 
a similar architecture was tested using a bi-directional LSTM (BLSTM) \cite{Hochreiter1997_LSTM}
model with attention mechanism (henceforth called BLSTMATTsim). 
The model takes log-Mel filterbank features 
as input and encodes them in a BLSTM with 2 layers of 512 nodes
followed by an attention mechanism which is then
projected down to the emotions under consideration. 
As initial experiments with this architecture did not
achieve the same results as published by \cite{Jalal2020_BLSTMATT}, 
a new model design was chosen, motivated by the consideration that a
combination of the strengths of CNNs (analysing spatial data) and
RNNs (analysing sequential data) could improve performance further.
The new model architecture feeds the same log-Mel filterbank features into a series of
CNN and RNN layers followed by a stack of fully connected layers and 
an attention mechanism (see Figure \ref{fig:CNNRNNmodel}
and henceforth called CNNRNNATT). The bi-directional LSTM network architecture 
is chosen as it considers a temporal feature distribution over the whole input sequence
to be useful for SER, as mentioned in \cite{Jalal2020_BLSTMATT}. 

For model training the machine learning toolkit 
LUDWIG\footnote{\label{1}\url{https://github.com/ludwig-ai/ludwig}}
was used which is built on the 
tensorflow library\footnote{\label{2}\url{https://www.tensorflow.org/}} \cite{Abadi2016_TENSORFLOW}
and uses python's SoundFile library\footnote{\label{3}\url{https://pypi.org/project/SoundFile}}
to read sound-files.

\subsection{Data augmentation}
\label{ssec:DataAug}
Data augmentation aims to improve model performance through 
an increase of the amount of training data which is typically done by altering
existing training material in the frequency or time domain (e.g.\ \cite{Snyder_2018_DatAug}). 
Often signal based methods are used where modifications such as speed or
volume perturbation are applied (e.g.\ in ASR \cite{Ko_2015_DatAug, Huang_2019_DatAug})
and new signals are generated, but there are also features based methods in which
features derived from signals are altered (e.g.\ \cite{Park_2019_SpecAugment}).
For this study, two data augmentation methods were tested: 1) speed and 2) volume perturbation. 

\vspace{-0.3cm}
\section{Corpora}
\label{sec:Corpora}
The following six speech corpora are used in the experiments:
IEMOCAP \cite{Busso2008_IEMOCAP}, RAVDESS \cite{RAVDESS_2018}, 
CMU-MOSEI \cite{zadeh2018_MOSEI} 
and three single speaker expressive speech corpora entitled TF1, TF2, and TM1
(‘F’ = female, ‘M’ = male speaker).

The first three corpora are publicly available corpora often used in emotion recognition research,
while the remaining three corpora are in-house expressive speech corpora. IEMOCAP, RAVDESS
and CMU-MOSEI provide multi-speaker recordings of which CMU-MOSEI is the only one to include
what is deemed to be natural emotions since it was recorded from YouTube videos.
IEMOCAP is considered to include `elicited' emotions which are induced by scripted and improvised
sessions between two actors (always male/female pairs).
RAVDESS includes acted emotions, which is also the case for the three single speaker corpora.
In terms of recording quality the three single speaker corpora are
high quality professional studio recordings, followed by RAVDESS,
IEMOCAP and CMU-MOSEI.

Corpus selection was based on covering a wide range of emotional expressions from
acted to natural and also the presence of the 4 emotion classes
{\it angry}, {\it happy}, {\it  sad} as well as {\it neutral},
which are the ones considered in this approach. Also, the presence of
single speaker corpora and multi-speaker corpora was intended enabling
a separate evaluation of combinations of them. 
In the following, each corpus is introduced in more detail and corpus specific splits into
train/test-sets are explained.

Table \ref{tab:NumUtts_in_6Corpora}
provides an overview of the 6 corpora along with their number of utterances
in total and by emotion. 

\begin{table}[htp!]
\small
\centering
\begin{tabular}{l|cccccc}         	\hline
 & IEM  & RAV  & MOS  & TF1    & TF2  & TM1\\ \hline
 Total     & 5531 &  672 &  6045 &  5430 & 5892 & 4383\\ \hline
{\it angry}    & 1103 &  192 &  316   & 764   & 627   & 627 \\
{\it happy}   & 1636  & 192 &  3757  & 761   & 1253 & 1253\\
{\it sad}       & 1084  & 192 &   686  & 754   &  1253 & 1252\\
{\it neutral}  & 1708  &  96  & 1286  & 3151  & 2759 & 1251\\ \hline
\end{tabular}
\caption{Number of utterances in total and by emotion class in each of the 6 corpora.}
\label{tab:NumUtts_in_6Corpora}
\end{table}

{\bf RAV: RAVDESS} (Ryerson Audio-Visual Database of Emotional Speech and Song) includes
North Amercian speech from 24 actors (12 male, 12 female) who recorded two sentences
in different emotions including {\it angry}, {\it happy}, {\it  sad} and {\it neutral}
(only speech was used, no song samples).
Speakers used two levels of intensity (normal=1, strong=2)
to realize emotions. Both levels were merged to indicate the presence of an emotion as opposed to absence.
To split the data into 5 folds for cross-validation always 19 speakers were used for training and 5 for testing
and the speakers were shifted in each fold. 

{\bf IEM: IEMOCAP} (Interactive Emotional Dyadic Motion Capture) \cite{Busso2008_IEMOCAP}
a widely used benchmark corpus in speech emotion recognition research; 
contains 10 speakers (5 male, 5 female) recorded in 5 sessions with either scripted or improvised
conversations in North American English. It provides about 12 hours
of audiovisual recordings of which only speech was used.  The corpus contains
more than 9 emotions annotated by multiple labelers.
It was recorded in a regular room with the microphone positioned between the two speakers,
therefore individual speech files associated with a single-speaker can still include audible parts
of the conversation partner in the background.
Many papers \cite{Milner2019_CROSSCORP,Jalal2020_BLSTMATT,Li2018_AttentionPooling}
conduct evaluations on IEMOCAP by selecting
a subset of 4 emotions: {\it angry}, {\it happy}, {\it neutral}, and {\it sad} and
build the {\it happy} class by combining 595 {\it happy} utterances and
1041 {\it excited} utterances, resulting in 1636 {\it happy} utterances; the same procedure
was applied in this study.
The 5 cross-validation sets were created with a leave-one-session-out
method, i.e.\ set 1 used sessions 1-4 for training and session 5 for testing. 

{\bf MOS: CMU-MOSEI} Carnegie Mellon University - Multimodal Opinion Sentiment and Emotion Intensity 
dataset \cite{zadeh2018_MOSEI} is a large corpus of more than 23.5k utterances
extracted from more than 1000 YouTube videos
which have been crowd-annotated with sentiment and emotion.
The language is English but a number of accents are present
and there is a wide variety of acoustic conditions including
different domestic background noises. Since the data was not specifically
designed for the purposes of emotion recognition, the emotions are considered to be natural.
Each utterance was annotated by 3 crowd-workers using a range of 0-3 for each emotion
in the set of {\it anger}, {\it disgust}, {\it fear}, {\it happiness}, {\it sadness}, and {\it surprise}.
For this study, only the emotions {\it angry}, {\it happy}, {\it sad} and {\it neutral}
(which is inferred by all emotions set to zero) are used.
In addition, since utterances can have multiple emotion labels,
only utterances for which there was a unequivocal annotation were chosen.
For creating 5 folds for cross-validation, the first fold was based on
the training, validation and test splits used for the
ACL 2018 conference\footnote{\label{4}\url{https://github.com/A2Zadeh/CMU-MultimodalSDK/blob/master/mmsdk/mmdatasdk/datasetstandard_datasets/CMU_MOSEI/cmu_mosei_std_folds.py}} and in other
folds split randomly but proportionally to the number of utterances per emotion class.

{\bf TF1,TF2,TM1}: The three in-house, single speaker expressive speech corpora 
include high quality studio recordings which were conducted with detailed
instructions for the speakers to deliver a consistent and distinctive speaking style for individual emotions.
TF1 includes natural speech recorded from a British English female voice talent,
TF2 provides expressive speech from a North American English female voice talent,
and TM1 contains expressive speech from a North American English male voice talent.
The data was split with a ratio of 80/20 into train/test-sets respectively and
in each set split randomly but proportionally to the number of utterances in each emotion class.

\begin{table}[htp!]
\small
\centering
\begin{tabular}{lcccccc}         	\hline
           & IEM   & RAV & MOS  & TF1  & TF2   & TM1\\ \hline
\#Train & 4290 & 912  & 4800 & 4329 & 4698 & 3494\\
\#Test  & 1241 & 240  & 1245 & 1103 & 1196 & 891\\ \hline
\end{tabular}
\caption{Number of utterances in the first fold of train/test sets.}
\label{tab:Utts_in_train-test-by-corp}
\end{table}

Table \ref{tab:Utts_in_train-test-by-corp} shows the number of utterances
used for training and testing for each of the six corpora as an example in the first fold. 

In addition to the above mentioned corpora, the
eNTERFACE \cite{Martin_2006_ENTERFACE} corpus was used as an example for a
completely unseen corpus to test various models on. The corpus includes
44 speakers (8 female) of which only 43 speakers were used because speaker 6 was unsegmented.
Each speaker produced 5 recordings of each emotion (acted) and only
{\it angry}, {\it happy}, {\it sad} were used as there are no {\it neutral}
recordings included. Speakers are speaking in different English accents.

\vspace{-0.3cm}
\subsection{Experimental set-up}
\label{ssec:ExperimentalSetup}

Using the CNNRNNATT model architecture, corpus-depen{\-}dent models were trained
and performance variations were observed between models
tested on matched test-sets (held-out data from the same corpus) and mismatched
test-sets (held-out data from other corpora). To avoid the influence of particular
splits into train/test-sets, 5-fold-cross validation was used for each corpus. This 
also enables the evaluation of performance variations across splits. 
Corpus specific aspects were taken into consideration to create
separate train/test-sets in the 5 folds as mentioned in
Section \ref{sec:Corpora} for each corpus.

After evaluating single corpus models,
corpora were combined and the impact on model performance measured. 
A model trained on all corpora was built and compared with the single corpus models
as well as with subsets of corpora.

To compare the impact of data augmentation methods, speed and volume
perturbation was used to augment corpora, effectively adding 2 copies of each corpus.
The combination of data complementation and data augmentation was also tested,
by training a model on the original data from all corpora plus the augmented data.  

For measuring the performance of models
the widely used unweighted accuracy (UA)
and weighted accuracy (WA) metrics were used,
as calculated by the following equations:

\begin{equation}
  UA = \frac{t_p + t_n}{t_p + t_n + f_p + f_n}
  \label{eq1}
\end{equation}

\begin{equation}
  WA = \frac{1}{2} (\frac{t_p}{t_p+f_n}+\frac{t_n}{t_n+f_p})
  \label{eq2}
\end{equation}

where $t_p =$ true positives, $t_n =$ true negatives,
$f_p =$ false positives, and $f_n =$ false negatives.
UA considers each class to have the same weight, while WA considers the
number of instances in each class to weigh its contribution.
Because there is a large number of comparisons which could not
all be fitted into the space restrictions in this article,
results are mainly reported in UA. 

\vspace{-0.3cm}
\section{EXPERIMENTS}
\label{sec:Experiments}
In an initial experiment the model architecture was established by 
first using a similar architecture as presented in \cite{Jalal2020_BLSTMATT}, i.e.\
the BLSTMATTsim model and then developing the new CNNRNN architecture
which are both introduced in \ref{ssec:ModelArchitecture}.

IEMOCAP \cite{Busso2008_IEMOCAP} was used to evaluate
model performance. To enable comparison with \cite{Jalal2020_BLSTMATT}
the identical split into train/test-sets was used, i.e.\ 4290 utterances (sessions 1-4) for training
and 1241 utterances (session 5) for testing.

Audio files were converted into a log Mel-spectral representation using
the `fbank' setting in LUDWIG. The parameters for the spectrum
extraction were: analysis window length = 26 ms,
window shift = 9 ms, number of filter bands = 23. 
The audio file length limit was set to 7.0 seconds which truncates files
longer than that and applies zero-padding to shorter files. 
Normalization was applied using the `per\_file' setting which
uses z-norm on a `per file' level. All audio-files had a sampling rate of 16 kHz. 

Results are summarized in Table \ref{tab:CompUA_CNNRNN_vs_baselines}
which shows three models from the literature above the divider
for comparison to our own models below the divider.
Results show that the BLSTMATTsim architecture resulted in lower performance than 
the figures reported in \cite{Jalal2020_BLSTMATT} for the BLSTMATT model.
The CNNRNNATT architecture achieved similar results as
\cite{Jalal2020_BLSTMATT}. The model named ``AttentionPooling''
from \cite{Li2018_AttentionPooling}
uses an attention pooling based representation learning method for SER. While,
\cite{Lian_2019_Att} uses both text and audio in a multimodal
model that combines attention modeling with a bi-directional
gated recurrent unit (GRU).
To get a more representative number for the CNNRNNATT model performance
across different train/test-splits of IEMOCAP, 5-fold cross validation with
leave-one-session-out splits was used and results are also shown
in Table \ref{tab:CompUA_CNNRNN_vs_baselines}, which serves as our baseline
for subsequent experiments. 
\begin{table}[htp!]
\small
\centering
\begin{tabular}{lcc}         	\hline
Model                        & UA   & WA\\ \hline
AttentionPooling \cite{Li2018_AttentionPooling} & 71.8 & --\\
Speech $+$ Text \cite{Lian_2019_Att} & 78.0 & --\\
BLSTMATT  \cite{Jalal2020_BLSTMATT}               & 80.1 & 73.5\\ \hline
BLSTMATTsim            & 76.5  & 68.7\\
CNNRNNATT    & 79.6 & 72.3 \\
CNNRNNATT [5fold]   & 76.4 & 68.6 \\ \hline

\end{tabular}
\caption{Comparison of chosen CNNRNNATT model architecture with previous models on the IEMOCAP 4-class data.}
\label{tab:CompUA_CNNRNN_vs_baselines}
\end{table}

The CNNRNNATT model takes the log-Mel filterbank features as input and
maps them to a tensor which is then passed through a stack of 6 convolutional
layers, followed by a stack of recurrent layers (here just 1),
and a stack of 4 fully connected layers
(sequence of nodes: 512-512-256-128)
with batch normalization after each layer and dropout set to 0.2. 
The reduce function was set to {\it null} which means that the full vector was output. 
The output layer is fed into the Bahdanau attention mechanism \cite{Bahdanau_2015}
which passes its output to the emotion classifier projecting to the 4 classes.
The loss function was selected as {\it sampled\_softmax\_cross\_entropy}.

To ensure a level playing field for the cross-corpus experiments all models were trained
with the same CNNRNNATT architecture using identical hyper-parameter settings.
Each model was trained for 200 epochs using the Adam optimiser \cite{Kingma2015_ADAM} 
with an initial learning rate of 0.0001 and a batch size of 186. Following \cite{Jalal2020_BLSTMATT}
the settings for reducing the learning rate when a plateau of validation measure is reached were 
4 epochs for patience with a reduction rate of 0.8.
\vspace{-0.2cm}
\section{Results}
\label{sec:Results}

\subsection{Data augmentation}
\label{ssec:DataAugResults}

To test the impact of data augmentation techniques, first, the IEMOCAP corpus was used.
For each augmentation technique one copy of the original
audio data was created by a) randomly altering the speed of the speech using 
the sox (v14.4.1, sox.sourceforge.net) {\it speed} effect
within a factor-range of $0.6$ and $1.5$ ($<1$ slows down, $>1$ speeds up), and 
b) randomly varying the volume within the same factor range using the sox {\it vol} effect.

Table \ref{tab:CompDataAug_IEM} shows the results in average UA
and average WA of 5-fold cross-validation
when applying data augmentation to IEMOCAP.
The combination of speech and volume perturbation was also tested
by generating 2 variants of each method and combining them with the original data
thereby effectively increasing the original training data five times  (shown in row entitled `2sp-2vol'). 

To check whether additional data created by other sox effects would result
in further performance improvements, the effects {\it bass} (boost/cut lower frequencies),
{\it treble} (boost/cut upper frequencies),
{\it overdrive} (non-linear distortion), and {\it tempo} (changing playback speed but not pitch)
were used to generate further variants which were used in the `7vars' model shown in
Table \ref{tab:CompDataAug_IEM} and increases the corpus size eight times. 

Results show that data augmentation with the mentioned effects can improve classification performance
and additive improvements are observed when combining effects as seen in the `2sp-2vol'
and `7vars' models. In addition, augmentation also seems to reduce the variance across
folds as can be seen by the reduced standard deviations in augmented models.
Not all effects are beneficial for performance boosting,
especially effects which cause more extreme signal distortions or
modulation factors which are too wide. 
It was also found that introducing more variation by random modulation factors
seemed to boost performance more then choosing fixed factors.   

\begin{table}[htp!]
\small
\centering
\begin{tabular}{lccc}         	\hline
Corpus         &  \#Train    & UA           & WA\\ \hline
IEM             & 4290        & 76.4 (1.9) & 68.6 (2.6)\\
$+$speed     & $\times2$  & 76.6 (1.2) & 68.9 (1.6)\\
$+$volume   & $\times2$  & 76.9 (1.5) & 69.2 (1.9)\\
$+$2sp-2vol & $\times5$  & 77.5 (1.5) & 70.1 (2.0)\\
$+$7vars     & $\times8$  & 78.2 (1.0) & 70.9 (1.4)\\ \hline
\end{tabular}
\caption{Comparison of data augmentation methods on IEMOCAP 4-class data. Results are average numbers for 5-fold cross-validation with standard deviations in brackets.}
\label{tab:CompDataAug_IEM}
\end{table}

\vspace{-0.3cm}
\subsection{Single corpus and cross-corpus models}
\label{ssec:SinglCorpCrossCorpResults}

Results for single corpus models and a model trained on all six corpora (All6)
are shown in Table \ref{tab:CrossCorpusUA_all_results_plus_Aug}.
In addition, Table \ref{tab:CrossCorpusUAaug_all_results_plus_Aug}
shows results when corpora are augmented with speed and volume perturbation, and 
when again, all six are combined and speed and volume augmentation is applied (All6aug).
The results for single corpus models
show that all of them perform best on matched test-sets and there are performance decreases
of different extends in mismatched conditions, i.e.\
generally, the multi-speaker, larger corpora IEM and MOS
show smaller performance reductions than the
single speaker corpora and the smaller, multi-speaker RAV corpus. 

Best performances are achieved by single speaker corpora with clean
studio recordings, acted emotions and
larger amounts of utterances for each emotion class (TF1, TF2, TM1),
followed by multi-speaker, acted emotions corpus RAV and
the lowest performances are computed on the
multi-speaker corpora MOS (natural emotions, variable recording conditions) and
IEM (elicited and acted emotions, normal room recordings with occasionally audible cross-talk).

The model trained on all 6 corpora (All6), as expected,
shows a significant improvement in overall performance
across all test-sets with an average UA of 87.5\%, while
it does not achieve the same level 
for the matched test-sets as seen in the
single corpus models - an indication that the
model fine-tunes very much onto a particular training corpus. 

Adding more training data by data augmentation (All6aug)
shows another, albeit small performance gain.

\begin{table*}[t]
\small
\centering
\begin{tabular}{ll|cccccc|c}         	\hline
 & & \multicolumn{7}{c}{\bf Model trained on}\\ \hline
 &  & IEM & RAV & MOS & TF1 & TF2 & TM1 & All6\\ \hline
\multirow{7}{*}{\begin{turn}{90}Tested on\end{turn}}
 & IEM & {\bf 76.4 (1.9)} & 65.0 (1.4) & 66.2 (1.5) & 67.7 (1.6) & 65.9 (1.1) & 63.8 (1.0) & {\bf 74.3 (0.6)}\\
 & RAV & 69.4 (1.1) & {\bf 85.9 (3.0)} & 64.4 (1.1) & 68.0 (1.7) & 68.1 (1.8) & 67.6 (2.7) & {\bf 78.4 (1.6)}\\
 & MOS & 68.7 (1.7) & 65.7 (3.0) & {\bf 77.4 (2.9)} & 60.5 (1.1) & 64.2 (0.8) & 64.2 (1.3) & {\bf 76.2 (1.2)}\\
 & TF1 & 63.8 (4.8) & 63.4 (1.1) & 71.1 (2.1) & {\bf 99.4 (0.2)} & 67.8 (0.3) & 59.5 (2.0) & {\bf 98.3 (0.3)}\\
 & TF2 & 60.1 (2.3) & 64.9 (1.2) & 66.2 (2.9) & 65.6 (0.4) & {\bf 99.9 (0.0)} & 60.9 (0.8) & {\bf 99.5 (0.1)}\\
 & TM1 & 60.8 (2.8) & 66.6 (1.5) & 68.0 (2.7) & 66.8 (0.7) & 65.3 (0.3) & {\bf 99.6 (0.4)} & {\bf 98.2 (0.3)}\\ \hline
 & Avg & 66.5 (6.1) & 68.6  (8.5) & 68.8 (4.7) & 71.3 (14.0) & 71.8 (13.8) & 69.3 (15.1) & 87.5  (12.3)\\ \hline
\end{tabular}
\caption{Average classification results in unweighted accuracy and standard deviation in brackets. Bold numbers show matched conditions. All6 is trained on all 6 corpora.}
\label{tab:CrossCorpusUA_all_results_plus_Aug}
\end{table*}
\begin{table*}[t]
\small
\centering
\begin{tabular}{ll|cccccc|c}         	\hline
 & & \multicolumn{7}{c}{\bf Model trained on}\\ \hline
 &  & IEMaug & RAVaug & MOSaug & TF1aug & TF2aug & TM1aug & All6aug\\ \hline
\multirow{7}{*}{\begin{turn}{90}Tested on\end{turn}}
& IEM & {\bf 76.5 (1.2)} & 66.0 (0.9) & 66.4 (2.0) & 69.5 (1.3) & 67.9 (0.8) & 65.4 (0.6) & {\bf 75.0 (0.7)}\\
& RAV & 71.3 (3.4) & {\bf 87.3 (1.5)} & 64.2 (1.3) & 70.6 (1.8) & 70.6 (1.1) & 68.4 (3.2) & {\bf 80.7 (2.2)}\\
& MOS & 68.2 (4.1) & 62.9 (2.7) & {\bf 77.8 (1.5)} & 61.5 (1.2) & 68.3 (1.9) & 65.2 (4.1) & {\bf 76.5 (3.0)}\\
& TF1 & 64.2 (5.5) & 65.6 (2.8) & 72.1 (2.0) & {\bf 99.4 (0.1)} & 72.2 (3.2) & 63.7 (1.2) & {\bf 98.3 (0.3)}\\
& TF2 & 61.5 (0.8) & 65.1 (2.7) & 68.7 (4.0) & 63.5 (0.4) & {\bf 99.9 (0.1)} & 61.9 (2.0) & {\bf 99.6 (0.2)}\\
& TM1 & 62.5 (3.0) & 67.0 (1.6) & 67.8 (1.4) & 72.0 (1.3) & 69.1 (1.6) & {\bf 99.5 (0.4)} & {\bf  98.4 (0.4)}\\ \hline
& Avg & 67.4 (5.8) & 69.0 (9.1) & 69.5 (4.8) & 72.8 (13.7) & 74.7 (12.5) & 70.7 (14.3) & 88.1 (11.9)\\ \hline
\end{tabular}
\caption{Average classification results in unweighted accuracy and standard deviation in brackets of models trained with augmented data (speed \& volume). Bold numbers are indicating matched conditions. All6aug is trained on all 6 corpora plus data augmentation for each corpus.}
\label{tab:CrossCorpusUAaug_all_results_plus_Aug}
\end{table*}

Looking at the distances in matched and mismatched conditions, 
average UA across all corpora for the 6 matched conditions is 89.8\%,
while for the 30 mismatched conditions it is 65.3\%. This distance is larger in single
speaker corpora than in multi-speaker corpora. 
When adding augmented data
the average UA across all corpora in matched conditions
improves slightly to 90.1\%, while the average UA across all mismatched
conditions improves to 66.8\%, i.e.\ a larger improvement
than in the matched case. An indication that performance of single corpus models
in mismatched conditions can be slightly boosted by data augmentation.

\begin{table}[t]
\small
\centering
\begin{tabular}{llccc}         	\hline
& & \multicolumn{3}{c}{\bf Model trained on} \\ \hline
& & IEMscript    & IEMimpro    & StudioScript\\ \hline
& & Train: 2078 & Train: 2212 & Train: 2004\\ \hline
 \multirow{4}{*}{\begin{turn}{90}Test.on\end{turn}}
& IEM        & 73.2 (1.6)  & 74.4 (1.1) & 68.6 (1.0)\\
& IEMscript & {\bf 77.4 (2.1)}  & 69.8 (0.7) & 66.2 (1.1)\\
& IEMimpro & 69.4 (3.3)  & {\bf 78.6 (1.5)} & 70.8 (1.2)\\ \hline
& Avg        & 73.3 (4.0)  & 74.3 (4.4) & 68.5 (2.3)\\ \hline
\end{tabular}
\caption{Average unweighted accuracy (standard deviation in brackets) for models trained
on IEMOCAP scripted (IEMscript) and improvised (IEMimpro)
and on a combination of RAV+TF1+TF2+TM1 for the StudioScript corpus. 
 Bold numbers are indicating matched conditions.}
\label{tab:ScriptVSImpoVSStudioRec_UA_all_results}
\end{table}

\begin{table}[t]
\small
\centering
\begin{tabular}{llcc}         	\hline
& & \multicolumn{2}{c}{\bf Model trained on} \\ \hline
& & IEMscript & IEMimpro\\ \hline
 & & Train: 2078 & Train: 2212\\ \hline
 \multirow{4}{*}{\begin{turn}{90}Tested on\end{turn}}
& RAV & 66.7 (1.6) & 68.1 (2.7)\\
& TF1 & 61.5 (4.2) & 66.7 (6.8)\\
& TF2 & 57.8 (1.1) & 59.8 (2.5) \\
& TM1 & 62.4 (3.4) & 59.1 (1.5)\\ \hline
& Avg & 62.1 (3.7) & 63.4 (4.6)\\ \hline
\end{tabular}
\caption{Average unweighted accuracy (standard deviation in brackets) for models trained
on IEMOCAP scripted (IEMscript) and improvised (IEMimpro) tested in mismatched conditions.}
\label{tab:ScriptVSImpro_mismatched_UA_all_results}
\end{table}

To investigate the influence of scripted vs.\ improvised subsets in IEMOCAP as well as
recording conditions in IEMOCAP vs.\ studio recordings another experiment was carried out.  
3 models are compared: IEMscript trained on the scripted utterances (2078 in training);
IEMimpro trained in the improvised sessions (2212 in training); and
StudioScript trained on a combination of scripted studio recordings
including utterance-balanced subsets of the 4 corpora RAV+TF1+\-TF2\-+TM1
with the following number of utterances by corpus: RAV=450, TF1=518, TF2=518, TM1=518.

Table \ref{tab:ScriptVSImpoVSStudioRec_UA_all_results} shows results when testing these
models on the full IEM test-sets and on the sub-test-sets from scripted and improvised sessions.
Table \ref{tab:ScriptVSImpro_mismatched_UA_all_results} shows results when evaluating
the IEMscript and IEMimpro models on mismatched test-sets. 
As can be seen the overall performance of IEMimpro is better than IEMscript on the
full IEM test-sets and also on mismatched test-sets from RAV, TF1, TF2 but not on TM1,
an indication that training on the improvised version provided better performance in both
matched and mismatched conditions. It is also interesting to observe
a speaker dependent influence when looking at the performance differences
between IEMscript and IEMimpro models on the mismatched single speaker corpora in
Table \ref{tab:ScriptVSImpro_mismatched_UA_all_results}:
IEMimpro works more than 5\% better on TF1 than IEMscript and 2\% better on TF2, but
on TM1 the IEMscript model is more than 3\% better than IEMimpro. 

To further test the performance on a completely unseen corpus both single corpus as well
as the multi-corpus model (All6) were tested on the unseen eNTERFACE corpus.
Results are shown in Table \ref{tab:TestonENTERFACE_UA_all_results} and
might be influenced by the special properties of eNTERFACE, i.e.\ reverberation
in recordings and multiple speakers speaking English in different accents.  
While the All6 model performs noticably better
than the single-corpus models MOS, TF1 and TM1, it is slightly below
IEM and TF2, and a bit less than 2\% below RAV. An indication
that just combining corpora does not solve the problem.

\begin{table*}[t]
\small
\centering
\begin{tabular}{lc|ccccccc}         	\hline
& & \multicolumn{7}{c}{\bf Model trained on}\\ \hline
  & ENT & RAV & IEM & MOS & TF1 & TF2 & TM1 & All6
\\ \hline
Test on ENT & 92.9 (1.8) & 70.1 (5.2) & 68.6 (6.3) & 62.7 (4.8) & 66.3 (9.9) & 68.7 (5.9) & 66.4 (4.9) & 68.4 (6.7)\\ \hline
\end{tabular}
\caption{Average unweighted accuracy (standard deviations in brackets) for models trained on one of 6 corpora and one model trained on all combined (All6) when tested on the unseen eNTERFACE (ENT) corpus. 
Model performance on eNTERFACE itself is shown in column ``ENT''.}
\label{tab:TestonENTERFACE_UA_all_results}
\end{table*}

\begin{table}[t]
\small
\centering
\begin{tabular}{ll|cccccc}         	\hline
 & &  \multicolumn{3}{c}{\bf Model trained on } \\ \hline
 &      & TF1+TF2 & TF1+TM1 & TF2+TM1\\ \hline
 \multirow{3}{*}{\begin{turn}{90}Test on\end{turn}}
 & TF1 & 99.2 (0.1) & 99.0 (0.5) & {\bf 77.0 (3.5)} \\
 & TF2 & 99.9 (0.1) & {\bf 68.6 (2.0)} & 99.8 (0.2) \\
 & TM1 & {\bf 71.1 (0.7)} & 98.6 (1.8) & 99.3 (0.5)\\ \hline
 & Avg & 90.1 (16.4)  & 88.7 (17.4) & 92.0 (13.0) \\ \hline
\end{tabular}
\caption{Average classification results in unweighted accuracy (UA) for models trained on
the 3 single speaker corpora. Each value is based on 5-fold cross validation
and numbers in brackets provide standard deviations. Bold numbers are indicating
mismatched conditions.}
\label{tab:SingleSpkrCorp_UA_all_results}
\end{table}

Since the 3 large single speaker corpora share the same recording conditions it 
 enables a more focused evaluation on the impacts of mixing corpora and testing
 in matched and mismatched conditions. These corpora were analyzed by training models on 
 all combination variants and testing in matched and mismatched conditions. Results of
 mixtures of corpora show that in mismatched conditions adding one corpus generally improved
 performance despite not reaching the same levels as in the matched conditions, e.g.\
 training on TF1+TF2 and testing on unseen speaker TM1 results in average UA of 71.1\%,
which is better than the single corpus mismatched cases of TF1 tested on TM1 with 66.8\%
and TF2 tested on TM1 with 65.3\%, but still far lower than the single corpus matched case
TM1 tested on TM1 with 99.6\%. 
Table \ref{tab:SingleSpkrCorp_UA_all_results} shows an overview of the 
results for training on the single speaker corpora and combinations thereof.

There could be a gender influence when looking at the mismatched results of the single speaker
corpora in which TM1, when tested on female speakers TF1 and TF2 performs 5-8\% lower
than testing the female speakers on the male speaker. However, TF1 actually performs slightly
better on TM1 than it does on TF2, indicating that other aspects, likely to do with the style in
which emotions are delivered, are at play here.

\section{CONCLUSIONS}
\label{sec:Conclusions}

A cross-corpus study for speech emotion recognition was presented to test
performances of deep learning models in matched (test-set from same corpus) and 
mismatched (test-set from different corpus) conditions. Six corpora including variations
along the dimensions of number of speakers, emotion type (acted, elicited, natural)
and acoustic scene (studio recording, regular room recording, variable recording conditions including
mixed background noise) were used in the study. Additionally, data augmentation was
applied in the form of speed and volume perturbation. 

The study has produced quantifiable evidence that the chosen model
fine-tunes to the training data and when confronted with unseen data (mismatched condition)
performs 10-40\% worse than for seen data (matched condition).

Single speaker corpora with acted emotions recorded in studios perform best showing
unweighted accuracy values up to 99.9\% while 
there is a noticeable performance decline in the range of 10-24\% for
multi-speaker corpora with more variable recording conditions
down to unweighted accuracy figures of 76.4\%. 

Recording conditions are also influential with clean studio recordings easier
to classify than regular office room recordings or recordings with mixed conditions
in aspects such as environments, background noise, etc.

On the question whether the combination of corpora improves or
degrades performance there is a mixed picture. 
The results showed that adding more data is generally
beneficial in mismatched conditions and does not significantly
decrease performance in matched conditions.
However, there are corpus dependent variations, e.g.\
improvements were observed for single speaker corpora
with added data from other single speaker corpora and when
tested on unseen single speaker corpora, but not always
when tested on multi-speaker corpora. 

Another conclusion is, that if the test speaker is not in the training data
then performance is significantly lower compared to models
trained on single speaker corpora which see the speaker
in training and test on held-out sets in testing.

Furthermore, experiments on IEMOCAP showed that
training on improvised data showed to be more beneficial
than training on scripted data even when tested on scripted data. 

However, a model trained on all six corpora achieved the highest
overall performance across all test-sets showing only small performance reductions
of 1-7\% compared to dedicated single corpus models. 

The results on data augmentation indicate that additional performance
gains can be achieved and especially single corpus models
in mismatched conditions seem to benefit most.
Performing data augmentation on the model trained on all corpora
resulted in the best overall performance across all test-sets,
showing that the combination of different corpora plus the
addition of augmented data proved to have an additive
effect on performance. 

The study has shown, that the challenge of handling unseen speakers
and different recording conditions in speech emotion recognition
is still unresolved. More investigations are required to understand
the variability of emotions and to create a model which is robust
to speakers and recording conditions.

\vfill
\pagebreak

\newpage
\bibliographystyle{IEEEbib}
\bibliography{paper_ASRU2021_for_ArXiv}

\end{document}